\documentclass[runningheads]{llncs}
\usepackage[T1]{fontenc}
\usepackage{graphicx}
\usepackage[table]{xcolor}
\usepackage{booktabs}    
\usepackage{multirow}    
\usepackage{graphicx}    
\usepackage{amsmath,amssymb,amsfonts}
\begin{document}
\title{Better Decomposition, Free Aggregation: A Synthesizer-Folding Framework for Multilingual Multi-Hop Question Answering}
\titlerunning{SYFER} 
%
\author{Yilin Wang\inst{1}\orcidID{0009-0009-2393-2330} \and
Yuchun Fan\inst{1} \and Weidong Bao\inst{1} \and Zili Wei\inst{1} \and Shi Feng\inst{1} \and Tong Xiao\inst{1} \and Zhengtao Yu\inst{2} \and Jingbo Zhu\inst{1}\textsuperscript{$\dagger$}}
\authorrunning{Wang et al.}
%
\institute{School of Computer Science and Engineering, Northeastern University, Shenyang, China \and Yunnan Key Laboratory of Artificial Intelligence, Kunming University of Science and Technology, Kunming, China
\\
\email{wangyilin0409@gmail.com}}
\maketitle              
\begin{abstract}
Multilingual retrieval-augmented generation (mRAG) equips large language models with access to globally distributed external knowledge for complex multilingual question answering. Recent approaches either translate retrieved documents into English or the query language to bridge the cross-lingual semantic gap, or decompose a complex query into sub-questions and aggregate the intermediate reasoning process. However, both lines of work suffer from two limitations. First, \textbf{one-size-fits-all translation alignment}, blanket translation discards culturally and linguistically native information unique to the target language, introduces translation noise, and inflates system cost. Second, \textbf{greedy decomposition and aggregation}, uncontrolled decomposition produces redundant sub-questions that compound errors during step-wise reasoning, and the final aggregation over reasoning paths further amplifies these errors. We address both with our method \textsc{Syfer}, a synthesizer-folding framework for multilingual multi-hop question answering that defers translation rather than applying it by default. \textsc{Syfer} first invokes a format-constrained decomposer to produce a sub-question graph in the original language, followed by a decomposition-quality check; when the check passes, sub-questions are answered sequentially under a retrieve-then-answer policy in the target language, and the English translation pathway with bilingual sub-question graph alignment is activated only when the check fails. Experiments across multiple languages show that \textsc{Syfer} attains competitive accuracy while striking a favourable balance between performance and computational cost. The code and data will be available at \url{https://github.com/f6ster/Syfer}

\end{abstract}

\section{Introduction}
\label{sec:introduction}

Retrieval-augmented generation (RAG) grounds large language models (LLMs) in external knowledge to improve factuality and mitigate hallucination on knowledge-intensive tasks~\cite{huang2025alleviating,huang2025improving,huang2025survey,lewis2020retrieval}.
However, in the multilingual setting, the training distribution of LLMs is highly skewed toward English and other high-resource languages, leaving a persistent capability gap on mid- and low-resource languages~\cite{fan2026lang,fan2025slam,huang2026survey,zhang2025multilingual}.
This motivates multilingual RAG (mRAG), which lets the model draw on multilingual collections to compensate for capability gaps and to recover information that is unevenly distributed across languages.

Existing mRAG systems handle single-hop multilingual queries well~\cite{chirkova2024retrieval,ranaldi2025improving} but struggle on multilingual multi-hop QA, where the LLM must also collect and integrate clues scattered across documents in different languages. The dominant remedy translates either the query or retrieved documents to align languages~\cite{park2025investigating,qi2025consistency,ranaldi2026multilingual}; a more refined line decomposes the query into a bilingual sub-question graph and aggregates per-hop answers~\cite{chen2026you,wang2026dapt,xiao2025lag}, which delivers stronger interpretability and currently represents one of the strongest design points on this task
\begin{figure*}[t]
  \centering
  \includegraphics[width=\textwidth]{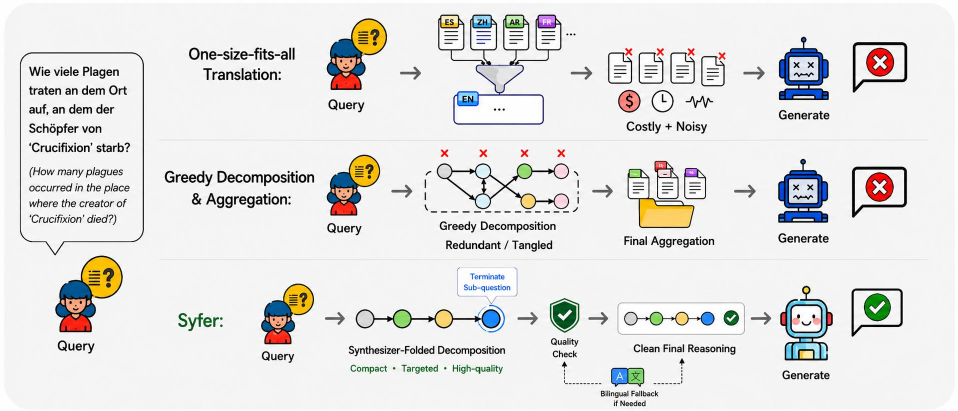}
  \vspace{-1.5em}
\caption{Two structural issues in prior decomposition-based mRAG and how \textsc{Syfer} addresses them. \emph{Top}: one-size-fits-all translation injects noise, and an end-of-pipeline aggregation call concentrates intermediate errors. \emph{Bottom}: \textsc{Syfer} reasons in the query language by default, falls back to a cross-lingual pathway only when faithfulness fails, and folds aggregation into a terminal sub-question.}
  \vspace{-1.5em}
  \label{fig:teaser}
\end{figure*}

Despite these advances, two structural issues remain, illustrated in Figure~\ref{fig:teaser}. (i) \textbf{One-size-fits-all translation}: forcibly aligning native-language documents to the query language incurs a heavy translation cost and tends to inject translation noise that is later answer as evidence. (ii) \textbf{Greedy decomposition and aggregation}: uncontrolled decomposition produces a sub-question graph that is often redundant and logically incoherent, where minor errors in intermediate reasoning propagate and compound. Besides, the final aggregation call is logically independent of the decomposition process, it concentrates rather than absorbs decomposition noise, breaking the reasoning chain and yielding incorrect answers.

We address both issues with \textsc{Syfer}, a \textbf{Sy}nthesizer-Folding \textbf{f}ram\textbf{e}wo\textbf{r}k for multilingual multi-hop question answering. \textsc{Syfer} keeps the decomposition-based backbone but makes two changes. First, translation becomes decomposition-driven: the framework reasons in the original language by default and activates the cross-lingual pathway only when the produced sub-question graph fails a quality check, after which an English-parallel graph is decomposed and fused for recovery. Second, decomposition becomes synthesizer-folded: a trained decomposer constrains the breadth and format of sub-questions and emits a terminal sub-question in the same logical layer as its peers. This terminal sub-question serves as a synthesis question over the prior reasoning chain, so the generator answers it directly instead of performing a separate aggregation over a long intermediate trace. As a result, aggregation is folded into decomposition and decomposition quality becomes directly checkable.

We further extend the multilingual multi-hop benchmarks introduced by~\cite{wang2026dapt} into a more comprehensive testbed spanning five language families and nine languages across high-, mid-, and low-resource regimes. Across three benchmarks and nine languages, \textsc{Syfer} not only attains competitive accuracy against strong baselines but also achieves a more favourable balance between performance, computational cost, and latency. The same experiments expose an interesting finding: unconditional fusion of an English-parallel sub-question graph into the reasoning trace is in fact sub-optimal on queries that decompose cleanly, since these queries already correspond to low-complexity reasoning, and a redundant English-side graph injects choices that confuse the generator rather than help it.
In summary, the main contributions of this paper are as follows:
\begin{itemize}
  \item We propose \textsc{Syfer}, a multilingual RAG framework that introduces synthesizer folding, the end-of-pipeline aggregation call is replaced by a terminal sub-question emitted by the decomposer in the same logical layer as its peers, placing decomposition and aggregation in a single logical pass and making decomposition quality directly verifiable.
  \item We extend three multilingual multi-hop QA benchmarks into a unified testbed covering five language families and nine languages across high-, mid- and low-resource regimes, providing a comprehensive evaluation suite for future multilingual RAG systems.
\item Across three benchmarks and nine languages, \textsc{Syfer} attains strong accuracy and a favourable cost-performance balance against competitive baselines. On the most challenging benchmark MuSiQue, \textsc{Syfer} still improves over the strongest decomposition-based baseline by $+8.91$ F1 ($+29.8\%$ relative) averaged over nine languages. 
\end{itemize}

\section{Related Work}
\label{sec:related-work}

\subsection{RAG Systems for Multi-hop QA}
\label{subsec:rw-multihop}

Multi-hop question answering requires a RAG system to retrieve and integrate evidence scattered across multiple documents. Existing methods mainly follow two directions: iterative retrieval that progressively refines supporting evidence~\cite{huang2026parammute,shao2023enhancing,trivedi2023interleaving,wei2026cirag}, and structured retrieval over trees or graphs that guides the LLM toward clue-dense documents or relevant context spans~\cite{chen2026you,edge2024local,gutierrez2025rag,xiao2025lag}. These methods, however, largely rely on the English-side understanding ability of the underlying LLM. In multilingual settings with parallel documents and mid/low-resource languages, such structures can introduce redundant evidence and amplify extraction noise, reducing downstream RAG effectiveness.

\subsection{Multilingual RAG}
\label{subsec:rw-mrag}

Multilingual RAG (mRAG) has primarily improved cross-lingual access through translation and alignment strategies~\cite{chirkova2024retrieval,park2025investigating,qi2025consistency,ranaldi2026multilingual,ranaldi2025improving,zhang2025multilingual}. These approaches typically translate the query into a high-resource pivot language or translate retrieved documents into the query language. While useful for language-aligned tasks, indiscriminate translation injects noise into the evidence and increases inference cost. Closer to our setting, DaPT~\cite{wang2026dapt} applies question decomposition to multilingual multi-hop QA, but its unconstrained decomposition produces long intermediate contexts and remains vulnerable to lost-in-the-middle effects~\cite{liu2024lost}.

In contrast, \textsc{Syfer} keeps reasoning in the original language by default and activates cross-lingual machinery only when the in-language decomposition fails a quality check. It further constrains the decomposition format and folds aggregation into a terminal sub-question, reducing both unnecessary translation and dependence on long intermediate reasoning traces.

\section{Methodology}
\label{sec:method}
\begin{figure}[t]
  \centering
  \vspace{-1em}
  \includegraphics[width=\linewidth]{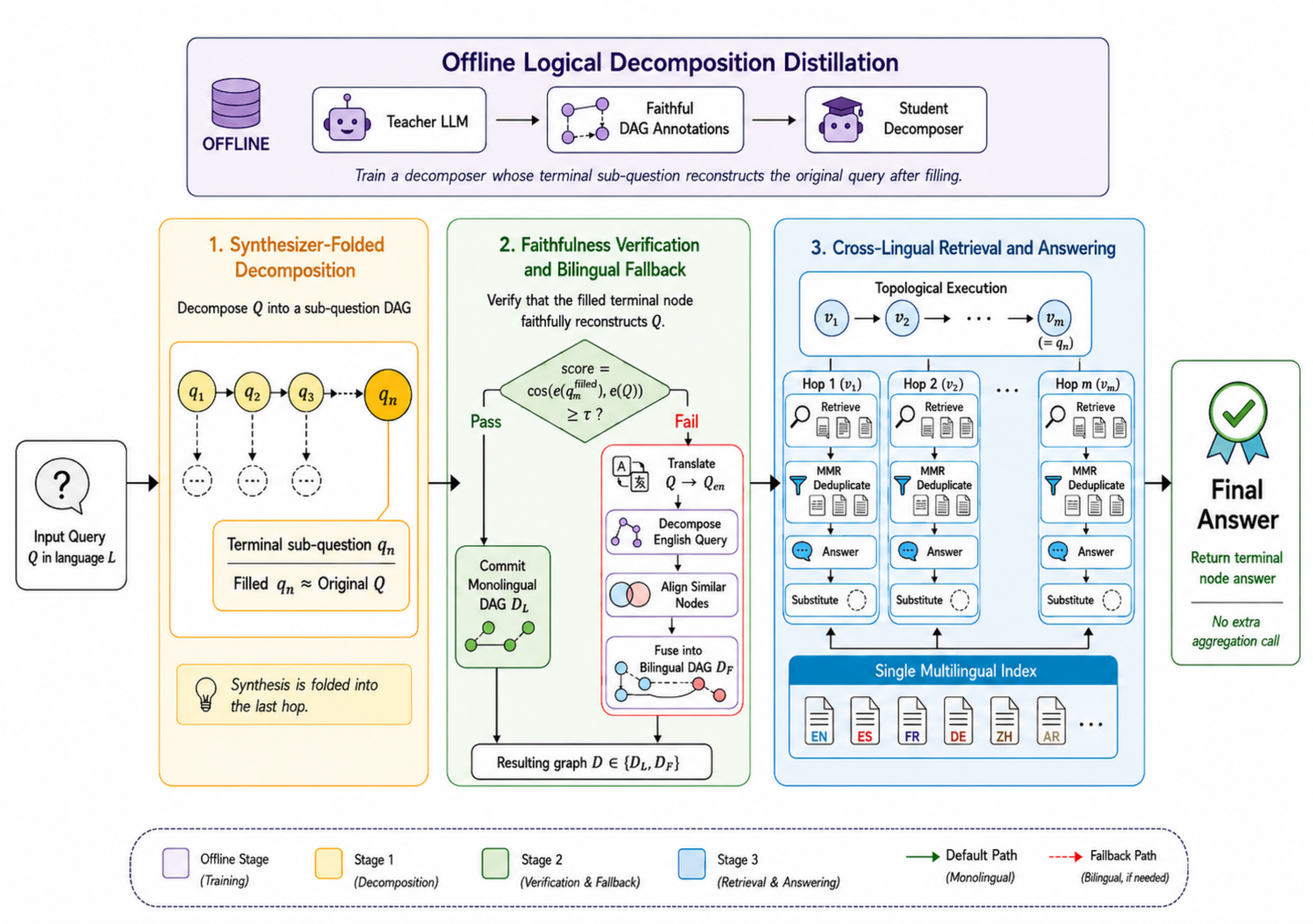}
  \vspace{-2em}
  \caption{Overview of \textsc{Syfer}: synthesizer-folded decomposition, faithfulness verification with bilingual fallback, and cross-lingual retrieval-and-answering.}
  \vspace{-1.5em}
  \label{f1}
\end{figure}

Given a query $Q$ in language $L$ and a multilingual corpus $\mathcal{C}$, \textsc{Syfer} retrieves evidence from $\mathcal{C}$ and uses the generator to produce the final answer $A$. As shown in Fig.~\ref{f1}, the framework consists of four stages: offline logical-decomposition distillation, synthesizer-folded decomposition, faithfulness verification with bilingual fallback, and cross-lingual retrieval-and-answering. The following subsections describe these stages in order.

\subsection{Logical Decomposition Distillation}
\label{ssec:distill}

A natural alternative is to invoke a frontier LLM to decompose every test query online, but this makes inference expensive and leaves decomposition quality hard to verify before answering. We instead distil a standalone decomposer on a corpus--query pool held out from evaluation, so that synthesizer folding is learned offline.

The teacher model maps each query $Q$ to an acyclic sub-question DAG $D^{*}$ with a unique terminal node and placeholder dependencies. We keep only decompositions whose filled terminal sub-question remains close to $Q$ in the retriever embedding space:
\begin{equation}
  \cos\!\left(\mathbf{e}(q_{n}^{\text{filled}}),\; \mathbf{e}(Q)\right) \;\geq\; \tau_{\text{constraint}},
  \label{eq:constraint}
\end{equation}
where $\mathbf{e}(\cdot)$ is the retriever embedding. This filtering yields the supervision set $\mathcal{D}_{\text{train}} = \{(Q^{(k)}, D^{*(k)})\}_{k=1}^{N}$.

Each $D^{*}$ is linearised into a token sequence $y = (y_{1}, \ldots, y_{T})$ that preserves node order and edge structure. The student decomposer with parameters $\theta$ is trained by token-level cross-entropy:

\begin{equation}
  \mathcal{L}_{\text{Distill}}(\theta)
  \;=\; -\,\mathbb{E}_{(Q,\,y)\sim \mathcal{D}_{\text{train}}}
  \sum_{t=1}^{T} \log p_{\theta}\!\left(y_{t} \,\middle|\, Q,\, y_{<t}\right),
  \quad
  \theta^{*} \;=\; \arg\min_{\theta}\;\mathcal{L}_{\text{Distill}}(\theta).
  \label{eq:sft-loss}
\end{equation}
The trained $\theta^{*}$ defines the inference-time decomposer in Section~\ref{ssec:syn-fold} and is reused, without branch-specific fine-tuning, by the bilingual fallback in Section~\ref{ssec:faith-fallback}.

\subsection{Synthesizer-Folded Decomposition}
\label{ssec:syn-fold}

Conventional multi-hop RAG ends with a dedicated aggregation call over $(Q, q_{1{:}n}, a_{1{:}n})$, where most cross-lingual noise accumulates. \textsc{Syfer} removes this call by letting the decomposer emit a terminal sub-question that, once filled, already speaks for $Q$. Concretely, given $Q$ in language $L$,
\begin{equation}
  D_{L} \;=\; \mathrm{Decompose}_{\theta^{*}}(Q),\qquad D_{L} = (V_{L}, E_{L}),
  \label{eq:decompose}
\end{equation}
with $V_{L} = \{q_{1}, \ldots, q_{n}\}$ and $E_{L} \subseteq V_{L} \times V_{L}$. An edge $(q_{i}, q_{j}) \in E_{L}$ indicates that the answer $a_{i}$ is required to instantiate $q_{j}$; substituting $a_{i}$ for each placeholder $\#i$ ($i<j$) in $q_{j}$ yields the filled form $q_{j}^{\text{filled}}$. The graph is acyclic and exposes a unique terminal node $q_{n}$. Trained under Eq.~\eqref{eq:constraint}, the terminal node is no longer an arbitrary leaf but a learned slot whose filled form encodes both $Q$ and the prior sub-answers, so that answering it against the corpus is, by construction, equivalent to aggregating.

\subsection{Faithfulness Verification and Bilingual Fallback}
\label{ssec:faith-fallback}

Even a well-trained decomposer is not fully reliable, for difficult queries, the filled terminal sub-question may drift away from the original query $Q$. \textsc{Syfer} therefore verifies the terminal sub-question before retrieval and uses bilingual fallback only when the monolingual decomposition appears unfaithful. Formally,
\begin{equation}
\begin{aligned}
  \mathrm{score}(D_{L})
  &= \cos\!\left(\mathbf{e}(q_{n}^{\text{filled}}),\, \mathbf{e}(Q)\right), \\
  \mathrm{Route}(D_{L})
  &=
  \begin{cases}
    D_{L}, & \mathrm{score}(D_{L}) \geq \tau_{\text{constraint}}, \\
    \mathrm{BilingualFallback}(D_{L}, Q),
    & \mathrm{score}(D_{L}) < \tau_{\text{constraint}}.
  \end{cases}
\end{aligned}
\label{eq:route}
\end{equation}
Thus, translation is not a default operation but a recovery path for queries whose in-language decomposition cannot be trusted.

When fallback is triggered, \textsc{Syfer} translates $Q$ into English, decomposes the English query with the same decomposer, and aligns the English DAG with the original-language DAG by node similarity:
\begin{equation}
  (q_{i}^{L*}, q_{j}^{\mathrm{en}*}) = \arg\max_{q_{i}^{L} \in V_{L},\, q_{j}^{\mathrm{en}} \in V_{\mathrm{en}}}
  \mathrm{sim}\!\left(\mathbf{e}(q_{i}^{L}),\, \mathbf{e}(q_{j}^{\mathrm{en}})\right).
  \label{eq:node-align}
\end{equation}
Node pairs above the alignment threshold $\tau_{\text{align}}$ are fused, yielding a bilingual graph
\begin{equation}
  D_{F} = \mathrm{Fuse}(D_{L}, D_{\mathrm{en}}),
  \label{eq:fused-graph}
\end{equation}

\subsection{Cross-Lingual Retrieval and Answering}
\label{ssec:retrieve-answer}

The output of the previous stage is either a monolingual graph $D_{L}$ or a fused bilingual graph $D_{F}$. \textsc{Syfer} solves either graph in topological order, so each sub-question is answered only after the sub-questions it depends on:
\begin{equation}
  S = (v_{1}, \ldots, v_{m}) = \mathrm{TopoSort}(D), \qquad D \in \{D_{L}, D_{F}\}.
  \label{eq:toposort}
\end{equation}
 For a monolingual node, \textsc{Syfer} retrieves with the target-language sub-question; for a bilingual node, it retrieves with both the target-language and English sub-questions and merges the candidates:
\begin{equation}
  R_{i} =
  \begin{cases}
    \mathrm{Retrieve}(q_{i}^{\text{filled}, L}, \mathcal{C}), & v_{i} \in V_{L}\setminus V_{\text{bi}}, \\
    \mathrm{Retrieve}(q_{i}^{\text{filled}, L}, \mathcal{C}) \cup \mathrm{Retrieve}(q_{j}^{\text{filled}, \mathrm{en}}, \mathcal{C}), & v_{i} \in V_{\text{bi}}.
  \end{cases}
  \label{eq:retrieval}
\end{equation}

Because the corpus contains document-aligned translations, the merged candidates may include several language versions of the same evidence. \textsc{Syfer} applies maximal marginal relevance (MMR) to keep documents that are both relevant to the current sub-question and different from documents already selected:
\begin{equation}
  d^{*} = \arg\max_{d \in R_{i} \setminus S_{i}}\;
  \lambda\, \mathrm{sim}(\mathbf{e}(q_{i}^{\text{filled}}), \mathbf{e}(d))
  \,-\, (1-\lambda)\, \max_{d' \in S_{i}}\, \mathrm{sim}(\mathbf{e}(d), \mathbf{e}(d')),
  \label{eq:mmr}
\end{equation}
where $S_{i}$ is the selected document set and $\lambda$ balances relevance and diversity. In plain terms, MMR prefers documents that match the sub-question while avoiding near-duplicate parallel translations.

Given the filtered evidence $R_i$, the generator produces one short answer per node, optionally using both language views for bilingual nodes:
\begin{equation}
  a_{i} =
  \begin{cases}
    \mathrm{Answer}(q_{i}^{\text{filled}, L}, R_{i}), & v_{i} \in V_{L}\setminus V_{\text{bi}}, \\
    \mathrm{Answer}(q_{i}^{\text{filled}, L}, q_{j}^{\text{filled}, \mathrm{en}}, R_{i}), & v_{i} \in V_{\text{bi}}.
  \end{cases}
  \label{eq:answer}
\end{equation}
Each answer is substituted into later sub-questions, and the answer to the terminal node is returned as the final answer:
\begin{equation}
  A = a_{m}.
  \label{eq:final-answer}
\end{equation}
This completes the synthesizer-folded pipeline without a separate final aggregation call.

\section{Experiments}
\label{sec:experiments}

\subsection{Datasets}
\label{ssec:datasets}
We evaluate \textsc{Syfer} on HotpotQA~\cite{yang2018hotpotqa}, 2WikiMultiHopQA (2Wiki)~\cite{ho2020constructing} and MuSiQue~\cite{trivedi2022musique}, using the same 1{,}000 query test split per benchmark as HippoRAG2~\cite{gutierrez2025rag}. To assess multilingual mRAG behaviour more comprehensively, we use GPT-4o~\cite{hurst2024gpt} to translate and extend the original English-only multi-hop test pool into nine languages spanning low, mid, and high-resource regimes across five language families. To ensure translation quality, we maintain a per-document entity table during translation so that entity names remain consistent across questions, supporting passages, and sub-questions.



\subsection{Baselines}
\label{ssec:baselines}

We compare \textsc{Syfer} with five representative baselines of multilingual multi-hop QA:
(i) \textit{Zero-shot LLM} and (ii) \textit{Vanilla RAG}, which answer without retrieval and with single-shot retrieval on the original query, respectively;
(iii) \textit{HippoRAG2}~\cite{gutierrez2025rag}, a strong structure-aware graph-RAG baseline operating over a triple-indexed knowledge graph;
(iv) \textit{CrossRAG}~\cite{ranaldi2026multilingual}, a retrieve-then-translate baseline that first retrieves documents on the multilingual corpus and then translates the retrieved documents into the query language before generating the response; and
(v) \textit{DaPT}~\cite{wang2026dapt}, a decomposition-based mRAG baseline that decomposes the query into a sub-question DAG and fuses each sub-question with its English-parallel counterpart at every hop.

\subsection{Metrics}
\label{ssec:metrics}

We report Exact Match (EM) and token-level F1 on the predicted answer string against the reference answer in the query language. EM assigns a score of $1$ only when the normalized prediction exactly matches the ground truth and $0$ otherwise. F1 measures token-level similarity by computing the harmonic mean of Precision and Recall, where Precision reflects the proportion of predicted tokens that are correct and Recall denotes the proportion of reference tokens successfully retrieved. Following MuSiQue~\cite{trivedi2022musique}, predictions and references are normalized with language-specific tokenization, lowercasing, and punctuation rules so that scores are comparable across the nine languages.

\subsection{Implementation Details}
\label{ssec:implementation}

\noindent\textbf{Backbones.}
To avoid model family preference, we use DeepSeek-V4 Pro~\cite{deepseekai2026deepseekv4} as the answering model for \textsc{Syfer} and for all baselines that require a generator. This decouples answer generation from both the decomposer family (Qwen) and the test-set translation model (GPT-4o), so that the three modeling layers do not share a common backbone and same-family confounders are ruled out in the cross-lingual evaluation.

\noindent\textbf{Retrieval setup.}
We use BGE-m3~\cite{chen2024bge} as the multilingual retriever for all methods, indexing the union of the nine-language corpora and retrieving $\text{top-}k = 5$ documents per query. The faithfulness gating threshold of \textsc{Syfer} is $\tau_{\text{constraint}} = 0.\text{8}$, the cross-lingual node alignment threshold is $\tau_{\text{align}} = 0.\text{6}$, and the MMR trade-off is $\lambda = 0.\text{6}$.

\noindent\textbf{Logical decomposition distillation.}
We sample a decomposer-training pool from the official training splits of the three benchmarks, disjoint from all test queries. Training-side translation and annotation are produced by Qwen3-235B-A22B-Instruct-2507~\cite{yang2025qwen3}, used as the teacher model. For \textsc{Syfer}, we curate 59{,}688 decomposition records covering six in-distribution languages: English (En), Chinese (Zh), German (De), Spanish (Es), Swahili (Sw), and Thai (Th). French (Fr), Bengali (Bn), and Korean (Ko) are held out as out-of-distribution evaluation languages. We fine-tune Qwen3-8B~\cite{yang2025qwen3} as the student decomposer for $2$ epochs with a global batch size of 64 and learning rate of $2{\times}10^{-4}$. Training is conducted on $8\times$NVIDIA H800 80\,GB GPUs with Intel(R) Xeon(R) Platinum 8468V CPUs.

\subsection{Main Results}
\label{ssec:main-results}
\begin{table*}[t]
\centering
\caption{EM and F1 scores across nine languages on the three multilingual multi-hop QA datasets we constructed. Avg. denotes the average score across languages. \textbf{Bold} values indicate the best performance within each group.}
\label{tab:main}

\definecolor{ourblue}{RGB}{220, 235, 250}
\resizebox{\textwidth}{!}{%
\begin{tabular}{l|*{10}{cc}}
\toprule
\multirow{2.5}{*}{\textbf{Method}}
 & \multicolumn{2}{c}{\textbf{En}} & \multicolumn{2}{c}{\textbf{Zh}}
 & \multicolumn{2}{c}{\textbf{De}} & \multicolumn{2}{c}{\textbf{Es}}
 & \multicolumn{2}{c}{\textbf{Sw}} & \multicolumn{2}{c}{\textbf{Th}}
 & \multicolumn{2}{c}{\textbf{Fr}} & \multicolumn{2}{c}{\textbf{Bn}}
 & \multicolumn{2}{c}{\textbf{Ko}} & \multicolumn{2}{c}{\textbf{Avg.}} \\
\cmidrule(lr){2-3}\cmidrule(lr){4-5}\cmidrule(lr){6-7}\cmidrule(lr){8-9}\cmidrule(lr){10-11}\cmidrule(lr){12-13}\cmidrule(lr){14-15}\cmidrule(lr){16-17}\cmidrule(lr){18-19}\cmidrule(lr){20-21}
 & EM & F1 & EM & F1 & EM & F1 & EM & F1 & EM & F1 & EM & F1 & EM & F1 & EM & F1 & EM & F1 & EM & F1 \\
\midrule
\multicolumn{21}{c}{\textbf{HotpotQA}} \\
\midrule
Zero-shot    & 18.6 & 28.2 & 10.2 & 27.9 & 13.1 & 21.0 & 14.3 & 23.6 &  8.8 & 15.4 & 13.2 & 40.0 & 15.1 & 22.7 &  8.2 & 14.0 &  7.7 & 23.7 & 12.1 & 24.1 \\
Vanilla RAG  & 35.4 & 48.4 & 24.8 & 42.9 & 25.4 & 38.8 & 22.1 & 35.3 & 26.4 & 41.1 & 22.6 & 44.2 & 15.1 & 27.3 &  8.4 & 21.2 & 12.3 & 33.1 & 21.4 & 36.9 \\
CrossRAG     & 34.6 & 47.5 & 13.4 & 30.2 & 24.6 & 37.5 & 22.3 & 35.3 & 25.8 & 40.5 & 21.5 & 41.4 & 16.3 & 27.7 &  5.3 & 18.2 &  8.9 & 26.9 & 19.2 & 33.9 \\
HippoRAG2    & 44.9 & 57.9 & 22.6 & 47.8 & 36.4 & 49.4 & 35.9 & 51.1 & 34.6 & 46.1 & 20.3 & 45.6 & 33.8 & 47.2 & 22.3 & 35.7 & 15.9 & 36.8 & 29.6 & 46.4 \\
DaPT         & 48.1 & 59.3 & 32.6 & 49.7 & 48.2 & 60.3 & 47.6 & 60.7 & 37.5 & 47.3 & 29.5 & 44.1 & 42.8 & 56.0 & 22.2 & 33.1 & 25.6 & 41.9 & 37.1 & 50.3 \\
\rowcolor{ourblue}
\textsc{Syfer} & \textbf{57.5} & \textbf{69.9} & \textbf{39.4} & \textbf{61.2} & \textbf{50.0} & \textbf{63.6} & \textbf{49.5} & \textbf{64.7} & \textbf{50.5} & \textbf{62.5} & \textbf{43.8} & \textbf{63.9} & \textbf{45.1} & \textbf{59.4} & \textbf{36.8} & \textbf{49.2} & \textbf{37.3} & \textbf{47.4} & \textbf{45.5} & \textbf{60.2} \\
\midrule
\multicolumn{21}{c}{\textbf{2WikiMultiHopQA}} \\
\midrule
Zero-shot    & 25.5 & 30.9 & 24.3 & 34.6 & 17.5 & 25.9 & 21.1 & 28.8 & 19.8 & 24.1 & 23.8 & 43.6 & 26.3 & 30.9 & 14.4 & 21.8 & 17.9 & 28.8 & 21.2 & 29.9 \\
Vanilla RAG  & 17.8 & 24.9 &  8.6 & 23.0 & 11.9 & 21.5 &  7.6 & 15.3 & 12.7 & 22.8 & 15.7 & 31.2 &  2.8 & 11.7 &  0.8 & 11.0 &  5.6 & 18.8 &  9.3 & 20.0 \\
CrossRAG     & 19.2 & 25.4 &  5.4 & 21.1 & 11.0 & 21.3 &  8.0 & 15.3 & 14.9 & 24.4 & 16.5 & 31.7 &  2.8 & 11.6 &  0.4 &  9.9 &  3.5 & 15.9 &  9.1 & 19.6 \\
HippoRAG2    & 56.0 & 65.0 & 21.5 & 45.0 & 45.2 & 56.1 & 44.4 & 57.1 & 40.3 & 49.5 & 32.4 & 52.1 & 42.5 & 52.9 & 22.5 & 32.7 & 20.8 & 36.8 & 36.2 & 49.7 \\
DaPT         & 34.7 & 39.3 & 30.0 & 45.0 & 43.9 & 50.9 & 43.1 & 50.1 & 32.7 & 36.5 & 26.2 & 34.7 & 41.3 & 50.2 & 24.3 & 34.2 & 29.7 & 41.0 & 34.0 & 42.4 \\
\rowcolor{ourblue}
\textsc{Syfer} & \textbf{67.5} & \textbf{75.2} & \textbf{42.8} & \textbf{61.7} & \textbf{64.2} & \textbf{72.4} & \textbf{64.6} & \textbf{73.2} & \textbf{60.4} & \textbf{68.3} & \textbf{57.3} & \textbf{71.3} & \textbf{59.0} & \textbf{68.1} & \textbf{45.1} & \textbf{53.7} & \textbf{45.9} & \textbf{59.2} & \textbf{56.3} & \textbf{67.0} \\
\midrule
\multicolumn{21}{c}{\textbf{MuSiQue}} \\
\midrule
Zero-shot    &  2.5 & 10.3 &  1.4 & 20.8 &  2.0 &  6.8 &  1.3 & 10.2 &  0.2 &  5.1 &  0.9 & 30.0 &  1.3 &  9.4 &  1.1 &  5.5 &  2.4 & 22.2 &  1.5 & 13.4 \\
Vanilla RAG  & 19.7 & 30.1 & 10.0 & 24.7 & 13.1 & 24.3 & 10.6 & 21.2 & 13.3 & 23.9 &  6.5 & 32.0 &  6.7 & 16.9 &  1.8 & 10.5 &  8.2 & 24.1 & 10.0 & 23.1 \\
CrossRAG     & 19.9 & 30.3 &  4.9 & 18.7 & 11.4 & 23.1 &  9.3 & 20.3 & 12.4 & 22.2 &  5.2 & 28.4 &  6.5 & 16.7 &  1.6 &  8.7 &  6.3 & 21.0 &  8.6 & 21.1 \\
HippoRAG2    & 24.3 & 35.9 &  7.8 & 30.9 & 17.2 & 26.3 & 17.4 & 31.2 & 14.9 & 25.5 &  4.5 & 35.6 & \textbf{16.2} & 29.1 &  7.6 & 17.8 &  8.0 & 28.6 & 13.1 & 29.0 \\
DaPT         & 28.1 & 35.6 & 19.1 & 34.0 & 24.8 & 34.5 & 24.6 & 36.0 & 16.7 & 23.9 &  9.1 & 27.0 & 19.8 & \textbf{31.6} &  9.4 & 18.3 & 12.6 & 28.1 & 18.2 & 29.9 \\
\rowcolor{ourblue}
\textsc{Syfer} & \textbf{40.9} & \textbf{51.4} & \textbf{22.4} & \textbf{40.7} & \textbf{30.1} & \textbf{41.6} & \textbf{29.1} & \textbf{42.2} & \textbf{30.5} & \textbf{39.9} & \textbf{20.0} & \textbf{47.1} & 11.0 & 19.9 & \textbf{15.8} & \textbf{27.1} & \textbf{20.0} & \textbf{39.4} & \textbf{24.4} & \textbf{38.8} \\
\bottomrule
\end{tabular}%
}

\vspace{-1.5em}
\end{table*}

Table~\ref{tab:main} reports EM and F1 of \textsc{Syfer} and the five baselines across nine languages on HotpotQA, 2WikiMultiHopQA, and MuSiQue. We organize the analysis around four observations.

\noindent\textbf{Single step retrieval is fundamentally insufficient for multi-hop QA, regardless of language alignment.} Vanilla RAG and CrossRAG both plateau at low EM/F1 across datasets and languages, and CrossRAG's retrieve-then-translate strategy fails to consistently improve over Vanilla RAG. The bottleneck is therefore not only cross-lingual mismatch but also one-shot retrieval's inability to gather all evidence for a compositional query, motivating decomposition-based pipelines.


\noindent\textbf{Structured graph indexing transfers poorly out of English.} 
HippoRAG2 is the strongest English baseline, but its advantage weakens on the multilingual corpus: from English to the nine-language average, its F1 drops by $23.6\%$ on 2Wiki ($65.0 \to 49.7$), $19.8\%$ on HotpotQA ($57.9 \to 46.4$), and $19.3\%$ on MuSiQue ($35.9 \to 29.0$). The drop is especially visible on Zh, Th, Bn, and Ko, suggesting that graph indexing based on LLM-extracted entities and relations is fragile when applied to heterogeneous multilingual evidence. The resulting cross-lingual noise and redundant edges then propagate into downstream retrieval.

\noindent\textbf{\textsc{Syfer} consistently enhances mRAG on multilingual multi-hop QA.} 
DaPT applies bilingual decomposition at every hop, so noisy sub-branches can still enter the final aggregation. \textsc{Syfer} instead gates bilingual reasoning by decomposition faithfulness (Eq.~\eqref{eq:route}) and returns the terminal sub-question answer directly (Eq.~\eqref{eq:constraint}), reducing the chance that unreliable branches dominate the final prediction. Empirically, \textsc{Syfer} achieves the largest average gain on 2Wiki, improving over the strongest baseline by $+17.3$ F1 and $+20.1$ EM, and remains robust on the harder MuSiQue benchmark with $+8.9$ F1 and $+6.2$ EM over DaPT. The same trend appears on HotpotQA and on the held-out OOD languages Fr, Bn and Ko, suggesting that the noise-suppression mechanism transfers beyond the languages used for decomposer training.

\begin{figure}[t]
\centering
\includegraphics[width=0.92\linewidth]{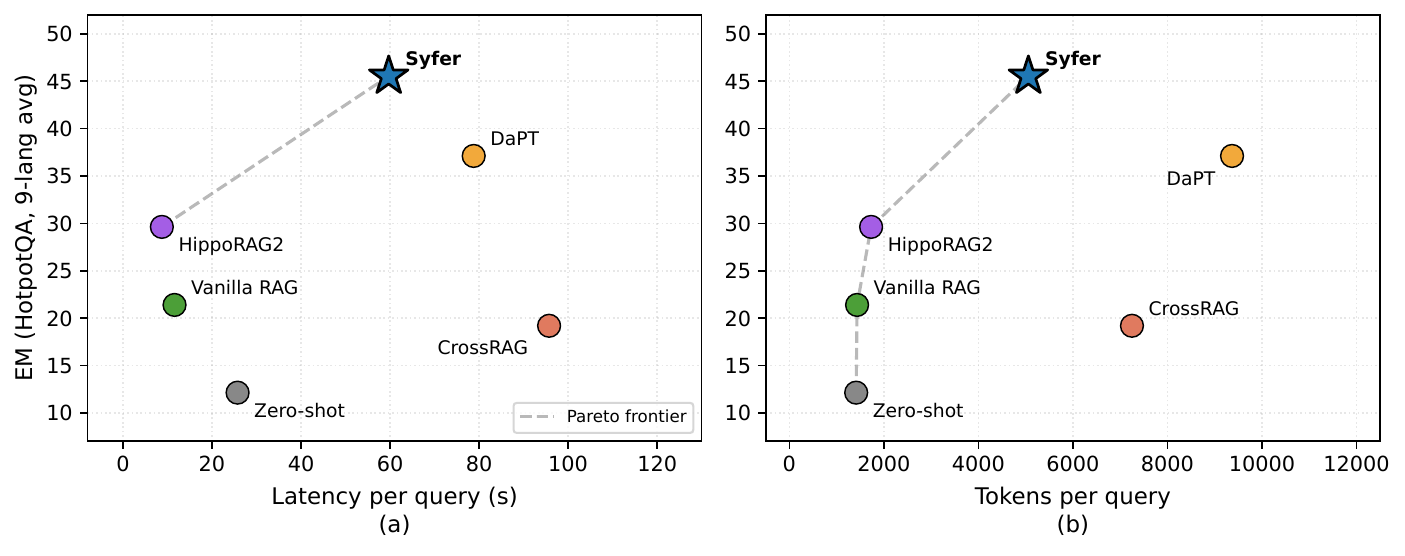}
\vspace{-1.5em}
\caption{Accuracy--cost Pareto fronts on the multilingual multi-hop QA pool. sub-figure (a): accuracy versus end-to-end latency per query. sub-figure (b): accuracy versus token cost per query. \textsc{Syfer} sits at the upper-right tip of the Pareto front in both panels.}
\label{fig:pareto}
\vspace{-1em}
\end{figure}

\noindent\textbf{\textsc{Syfer} attains a more favourable accuracy--cost balance than competing baselines.}
Figure~\ref{fig:pareto} plots the accuracy--cost Pareto fronts of \textsc{Syfer} against the five baselines, and \textsc{Syfer} achieves a better balance between accuracy and inference cost than every competitor. Notably, HippoRAG2 additionally requires hours of offline graph-index construction before any query can be served, an overhead that grows further as the corpus size increases.

\subsection{Ablation Study}
\label{ssec:ablation}
\begin{table*}[t]
\centering
\caption{Ablation performance on multilingual HotpotQA. \textbf{Bold} values indicate the best performance within each group.}
\label{tab:ablation}

\definecolor{ourblue}{RGB}{220, 235, 250}
\resizebox{\textwidth}{!}{%
\begin{tabular}{l|*{9}{cc}}
\toprule
\multirow{2.5}{*}{\textbf{Variant}}
 & \multicolumn{2}{c}{\textbf{Zh}} & \multicolumn{2}{c}{\textbf{De}} & \multicolumn{2}{c}{\textbf{Es}}
 & \multicolumn{2}{c}{\textbf{Sw}} & \multicolumn{2}{c}{\textbf{Th}}
 & \multicolumn{2}{c}{\textbf{Fr}} & \multicolumn{2}{c}{\textbf{Bn}}
 & \multicolumn{2}{c}{\textbf{Ko}} & \multicolumn{2}{c}{\textbf{Avg.}} \\
\cmidrule(lr){2-3}\cmidrule(lr){4-5}\cmidrule(lr){6-7}\cmidrule(lr){8-9}\cmidrule(lr){10-11}\cmidrule(lr){12-13}\cmidrule(lr){14-15}\cmidrule(lr){16-17}\cmidrule(lr){18-19}
 & EM & F1 & EM & F1 & EM & F1 & EM & F1 & EM & F1 & EM & F1 & EM & F1 & EM & F1 & EM & F1 \\
\midrule
\rowcolor{ourblue}
Full \textsc{Syfer} & \textbf{39.4} & \textbf{61.2} & \textbf{50.0} & \textbf{63.6} & 49.5 & \textbf{64.7} & \textbf{50.5} & \textbf{62.5} & \textbf{43.8} & \textbf{63.9} & \textbf{45.1} & \textbf{59.4} & \textbf{36.8} & \textbf{49.2} & \textbf{37.3} & \textbf{47.4} & \textbf{44.1} & \textbf{59.0} \\
\midrule
w/o Folding      & 33.2 & 52.0 & 49.5 & 62.7 & \textbf{50.8} & 63.2 & 35.9 & 44.4 & 31.6 & 47.4 & 44.0 & 56.9 & 26.9 & 38.9 & 30.2 & 46.2 & 37.8 & 51.4 \\
w/o Verification & 33.0 & 50.5 & 48.8 & 60.9 & 47.9 & 60.8 & 32.5 & 41.0 & 28.7 & 42.9 & 42.4 & 56.8 & 22.4 & 33.3 & 25.8 & 42.1 & 35.2 & 48.5 \\
w/o MMR          & 35.9 & 52.8 & 40.5 & 50.6 & 42.9 & 55.0 & 27.1 & 34.2 & 26.6 & 41.5 & 35.2 & 48.0 & 22.6 & 34.4 & 26.1 & 43.9 & 32.1 & 45.1 \\
Always Bilingual & 31.2 & 47.4 & 43.5 & 54.1 & 44.1 & 56.2 & 29.8 & 37.7 & 27.8 & 43.2 & 36.3 & 49.3 & 17.9 & 28.8 & 22.8 & 37.8 & 31.7 & 44.3 \\
\bottomrule
\end{tabular}%
}

\vspace{-0.5em}
\end{table*}
We ablate the four core components of \textsc{Syfer} on the eight-language multilingual test pool and report the average EM/F1 in Table~\ref{tab:ablation}. The ablated variants are:
(i) \textit{w/o Folding}, which restores the classical end-of-pipeline aggregation call;
(ii) \textit{w/o Verification}, which always commits the monolingual branch, never triggering the bilingual fallback;
(iii) \textit{Always Bilingual}, which keeps synthesizer folding but disables the faithfulness gate, running the bilingual branch at every hop unconditionally;
(iv) \textit{w/o MMR}, which replaces cross-lingual MMR with vanilla top-$k$.

Removing any component consistently weakens \textsc{Syfer}, confirming that the gains come from the joint design. Among the components, cross-lingual MMR and faithfulness-controlled bilingual routing are especially important: without them, retrieval is more easily dominated by near-duplicate parallel passages or by unnecessary bilingual branches. There is an interesting observation, \textit{Always Bilingual}, which confirms that more cross-lingual signal is not always better. This suggests that when a sub-question is already answerable in the target language, forcing an English-parallel branch injects extra reasoning noise and may even make the final answer drift into English, which is penalized by the language-conditioned EM/F1 metric.

\section{Conclusion}
\label{sec:conclusion}


We present \textsc{Syfer}, a synthesizer-folding mRAG framework for multilingual multi-hop QA that gates bilingual fusion on a faithfulness check and folds aggregation into a terminal sub-question, making decomposition quality directly verifiable. On a unified nine-language testbed extended from three multi-hop QA benchmarks, \textsc{Syfer} outperforms structure-aware, translation-based, and decomposition-based baselines with a better quality--cost trade-off.

\section{Acknowledgements}
\label{sec:conclusion}
This work was supported in part by the National Science Foundation of China (Nos. 62276056 and U24A20334), the Natural Science Foundation of Liaoning Province of China (2022-KF-26-01), the Fundamental Research Funds for the Central Universities (Nos. N2216016 and N2316002), the Yunnan Fundamental Research Projects (No. 202401BC070021), and the Program of Introducing Talents of Discipline to Universities, Plan 111 (No.B16009).

%
%
%
\bibliographystyle{splncs04}
\bibliography{references}
%




\end{document}